\documentclass{article}
\usepackage{spconf,amsmath,graphicx}
\usepackage{hyperref}
\usepackage{algorithmic}
\usepackage{subfigure}
\usepackage[font=small,skip=3pt]{caption}
\usepackage[linesnumbered,ruled,vlined]{algorithm2e}
\SetKwInput{KwInput}{Input}                
\SetKwInput{KwOutput}{Output}              

\title{Radar+RGB Attentive Fusion for Robust Object Detection in Autonomous Vehicles}
\name{Ritu Yadav, Axel Vierling, Karsten Berns}
\address{Technische Universität Kaiserslautern, Germany}
\begin{document}
%
\maketitle
\begin{abstract}
This paper presents two variations of architecture referred to as RANet and BIRANet. The proposed architecture aims to use radar signal data along with RGB camera images to form a robust detection network that works efficiently, even in variable lighting and weather conditions such as rain, dust, fog, and others. First, radar information is fused in the feature extractor network. Second, radar points are used to generate guided anchors. Third, a method is proposed to improve region proposal network~\cite{ren2015faster} targets. BIRANet yields 72.3/75.3\% average AP/AR on the NuScenes~\cite{caesar2019nuscenes} dataset, which is better than the performance of our base network Faster-RCNN with Feature pyramid network(FFPN)~\cite{lin2017feature}. RANet gives 69.6/71.9\% average AP/AR on the same dataset, which is reasonably acceptable performance. Also, both BIRANet and RANet are evaluated to be robust towards the noise.
\end{abstract}
\section{Introduction}
\label{sec:intro}

With increasing demand and research in autonomous vehicles, object detection has become a significant task. RGB images through cameras are the standard input of such object detectors. However, bad weather conditions or dark environment conditions distort the images or videos captured by a camera in a driving scenario. Unpredictable environmental conditions make it troublesome for single modality based object detectors to detect the object's location, and it is category precisely. To overcome this limitation, Vehicle industries are using different sensors in autonomous vehicles such as LIDAR, radar, infrared cameras, and others. Each sensor has it is own benefits and drawbacks. However, LIDAR and radar sensors have most commonly come into the picture when analyzed on the usability factor. LIDAR sensors are highly efficient in making 3d images of objects but at the cost of the reliability issue. LIDAR has more moving parts hence have room for more noise. Radar sensors, on the other hand, provide very few points hence are not suitable for 3d object image construction. However, radar sensors have a high range, due to which they can detect objects at a long distance. They are robust, reliable, and cheaper than LIDAR sensors. Long-range object detection is beneficial for autonomous vehicles, especially commercial vehicles such as trucks, where they need more time to slow down for an incoming obstacle or object. Radar sensors and RGB cameras, make a very reliable combination for object detection. Even after these significant merits of radar sensors, there is limited significant research work conducted on radar sensor and vision fusion.

In this paper, an architecture for object detection is introduced, which utilizes RGB camera images and radar sensor data. Radar points store location information of objects and their distance from the sensor with high confidence. We use this high confidence radar sensor information for better feature extraction and region proposals along with other intuitive ideas. Proposed approaches are implemented to build a robust object detection network for the autonomous driving environment. For the demonstration of the effectiveness of the proposed architecture, the NuScenes dataset is used, which has recorded data from a full autonomous sensor suite(6 RGB cameras, 1 LIDAR, 5 RADAR, GPS, IMU).

\section{Related Work}
\label{sec:format}
In autonomous vehicles, radar-based object detection is explored in work such as~\cite{chavez2012frontal, duan2016moving, kato2002obstacle, manjunath2018radar}. All these works are based on traditional techniques such as Haar-like features~\cite{neumann2017online}, Extended Kalman Filter(EKF)~\cite{chavez2012frontal}, occupancy grid map~\cite{duan2016moving}, and others.
With the introduction of Convolutional Neural Network(CNN), an informative representation of objects is learned, and it shows drastic improvement in tasks such as classification and regression. CNN based object detectors are very efficient and common these days. Two-stage object detectors generally overpower single-stage detectors in terms of accuracy. The main strength of the two-stage object detector is the region proposal network. Faster-RCNN~\cite{ren2015faster} introduced region proposal network(RPN) where they made network to learn anchor proposals. RPN was a tremendous success in the field of object detection. RPN is further explored in multiple works such as GA-RPN~\cite{wang2019region}, Iterative RPN~\cite{gidaris2016attend}, Cascade RPN~\cite{vu2019cascade}, and others. 

RPN is much explored on RGB camera images and very less on radar data. However, there are few works such as ~\cite{meyer2019deep}, where authors presented a 3D object detection network for car detection using AVOD architecture on radar pointclouds and camera images. Authors in ~\cite{zhang2019object}, use range-doppler spectrums from FMCW radar signals and camera images for 3D object detection and estimation.

Authors in ~\cite{nabati2019rrpn}, presented radar points-based anchors where radar points are considered towards the center, left, right, and bottom of the object. However, the authors did not provide any specific reason for not considering the radar point at the top of the object. The idea of radar point-based anchor generation is accommodated in the presented work with some intuitive changes.

\section{Proposed Object Detection Network}
\label{sec:format}
The idea of sensor fusion is inspired by humans. We humans use multiple sensors to sense different attributes or specifications, which are refined together to come to a final observation. When it comes to artificial sensor fusion, the same approach is followed. In the proposed architecture, input from RGB cameras and radar sensors are fused, refined, and processed for object detection.
Two variations of networks are proposed in this work, namely RANet and BIRANet. Both follow the same architecture with two differences; First, RANet works only on radar point-based anchors. However, BIRANet also utilizes anchors generated similar to that in FFPN. Second, BIRANet has a different RPN target generation method, "Best of Two," which is explained in subsection \ref{subsec:B2}. Below is a detailed description of the proposed architecture.
\subsection{Radar + RGB Fusion}
Radar sensors provide points, distance, velocity, azimuth, and others. Radar points are in the 3D coordinate system(x,y,z) where x and y coordinate of radar points provide the 2D location of reflective objects in the view, and the third dimension z corresponds to the distance of the object from the radar sensor.
\begin{figure}[htbp]
\centerline{\includegraphics[width=60mm]{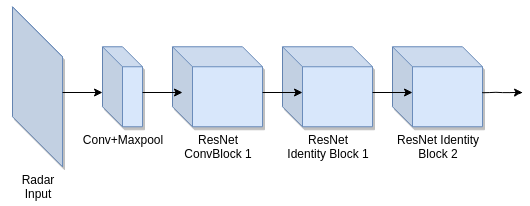}}
\caption{Radar Feature extractor branch.}
\label{fig1}
\end{figure}
Before feeding the radar data into the network, radar data points are first converted from vehicle coordinates to camera-view coordinates, which will align radar points with the RGB image. Second, radar points are resized and then converted into a feature map of a size equivalent to that of the input image. At each radar point location x, y of the feature map, the z distance value is embedded and assign 0 to all other locations where there is no radar point present. The resulting radar feature map is the input of radar feature extractor and has one channel.

\begin{figure}[htbp]
\centerline{\includegraphics[width=80mm]{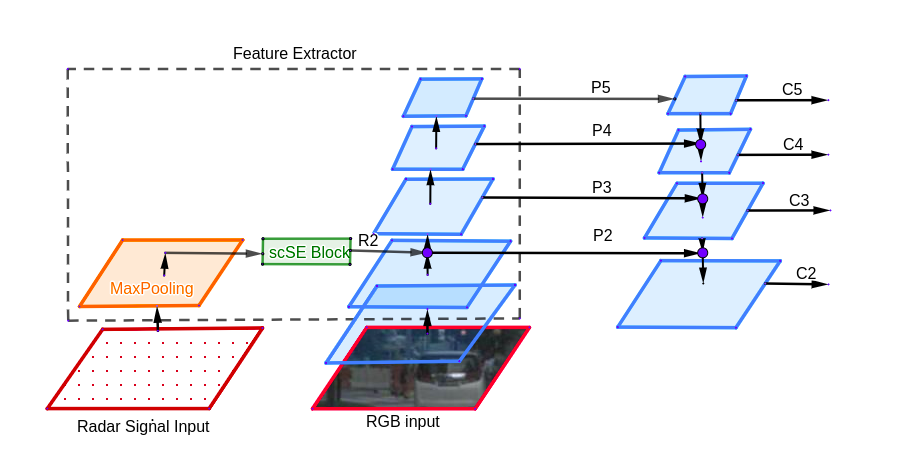}}
\caption{Radar+RGB featuremap fusion. Small purple circles represents element-wise addition. Red dots in radar signal input represent radar points.}
\label{fig2}
\end{figure}
In the proposed architecture, a separate radar feature extractor branch is formed, as represented in Fig \ref{fig1}. The idea of residual network is adapted to extract radar features efficiently, and one convolution along with two identity blocks is applied in the radar feature extractor branch. The radar feature extractor branch and RGB feature extractor branch are fused in the feature extractor backbone. The feature extractor used for RGB images is ResNet~\cite{he2016deep}. While we experimented with radar and RGB data fusion at different stages of the feature extractor, the proposed configuration shown in Fig \ref{fig2} proved to be the best. For fusion, element-wise addition operation is performed on the output feature map of the radar feature extractor branch and second stage output feature map of the RGB feature extractor branch.
\subsection{Attentive FPN}
In Feature pyramid network(FPN) of FFPN 4 feature map output, P2 to P5 are extracted from the feature extractor backbone. Extracted feature maps are the features on which the performance of the detection network is primarily based. Hence these features have to be as good as possible. 
It is experimented and found that in the radar+RGB fused network, the output feature maps can be boosted further with the help of attention. With this motivation, an Attentive FPN is proposed, which uses Concurrent Spatial and Channel ‘Squeeze \& Excitation’(scSE) blocks~\cite{roy2018concurrent}. The scSE block acts as attention and highlights important spatial features as well as significant channels. In the attentive FPN, scSE attention block is applied to the radar feature map. scSE blocks adaptively boost activation of areas where we have radar points and suppress the activation at other locations. Boosted good feature maps are then fused with RGB features map as shown in Fig \ref{fig2}.

\begin{figure}
    \centering
    \subfigure{\includegraphics[width=0.09\textwidth]{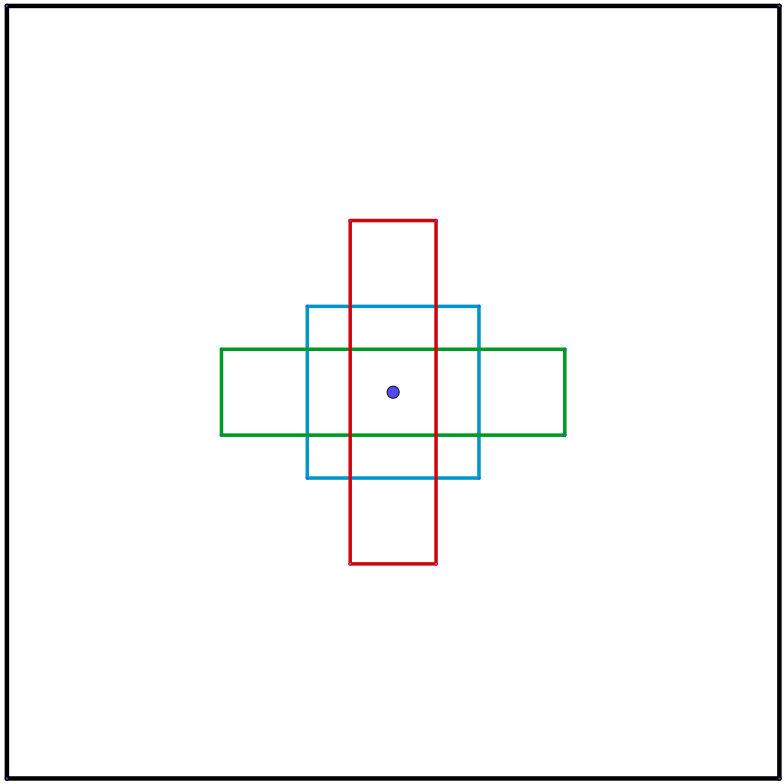}} 
    \subfigure{\includegraphics[width=0.09\textwidth]{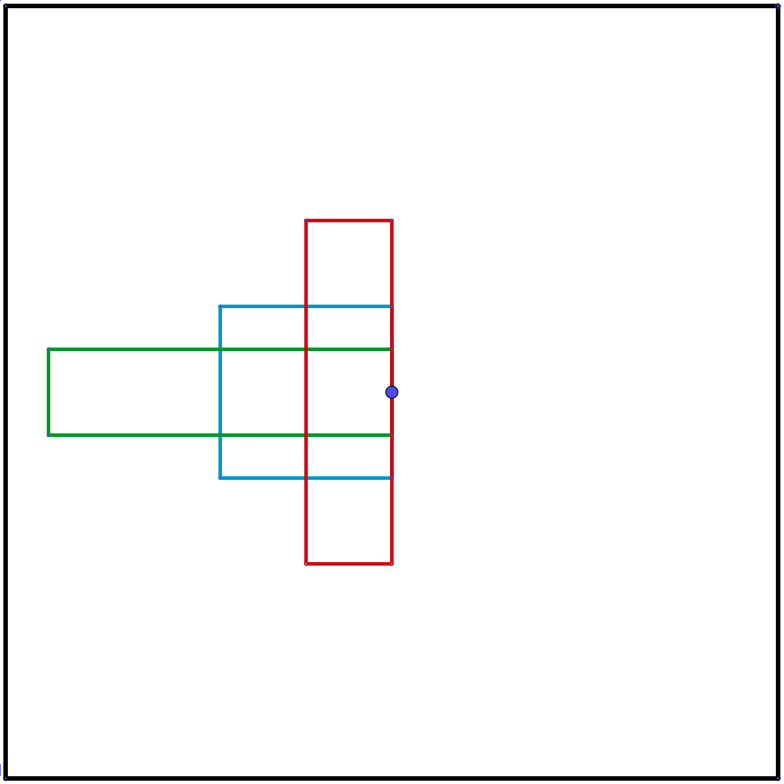}} 
    \subfigure{\includegraphics[width=0.09\textwidth]{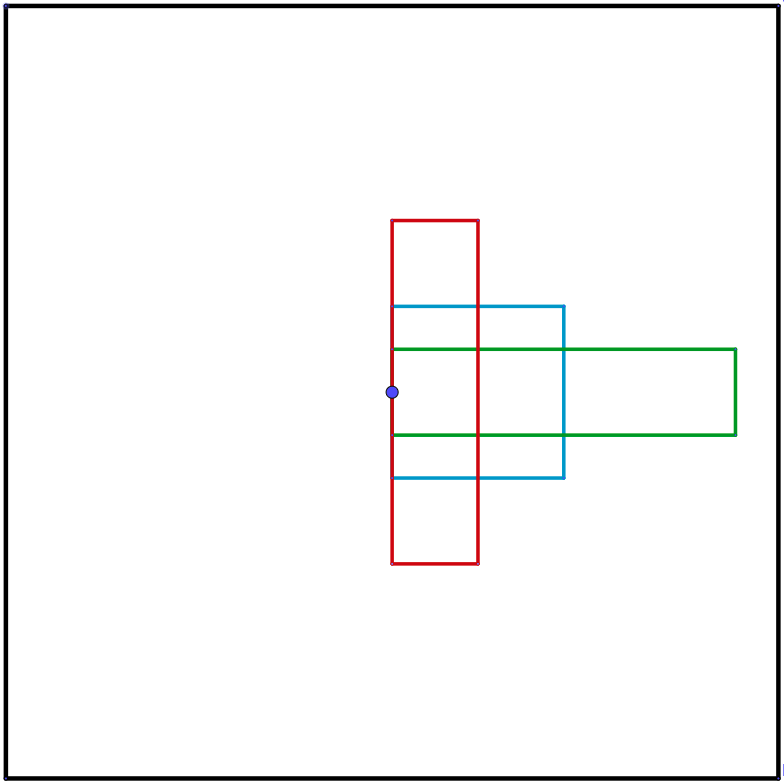}}
    \subfigure{\includegraphics[width=0.09\textwidth]{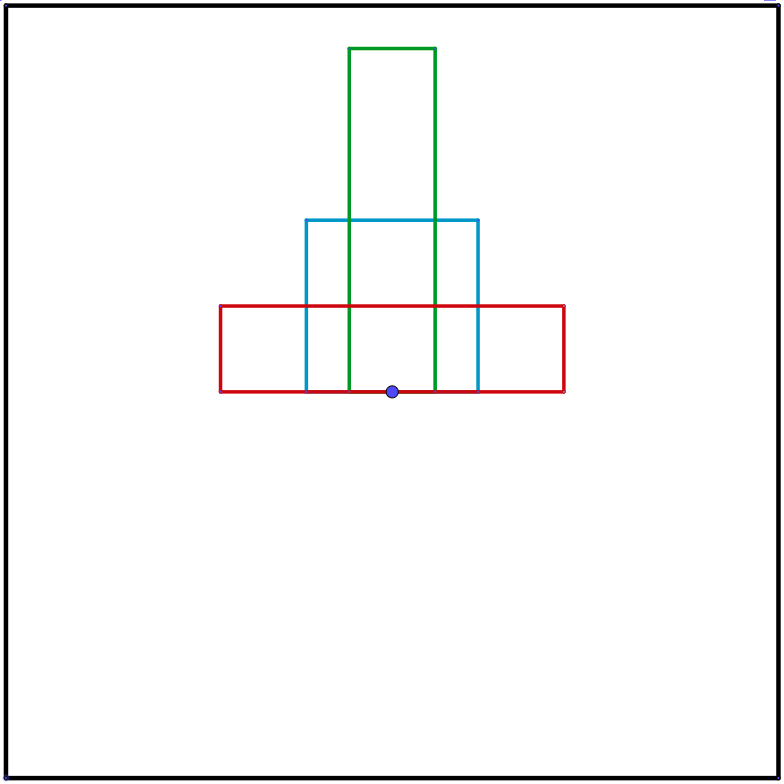}}
    \subfigure{\includegraphics[width=0.09\textwidth]{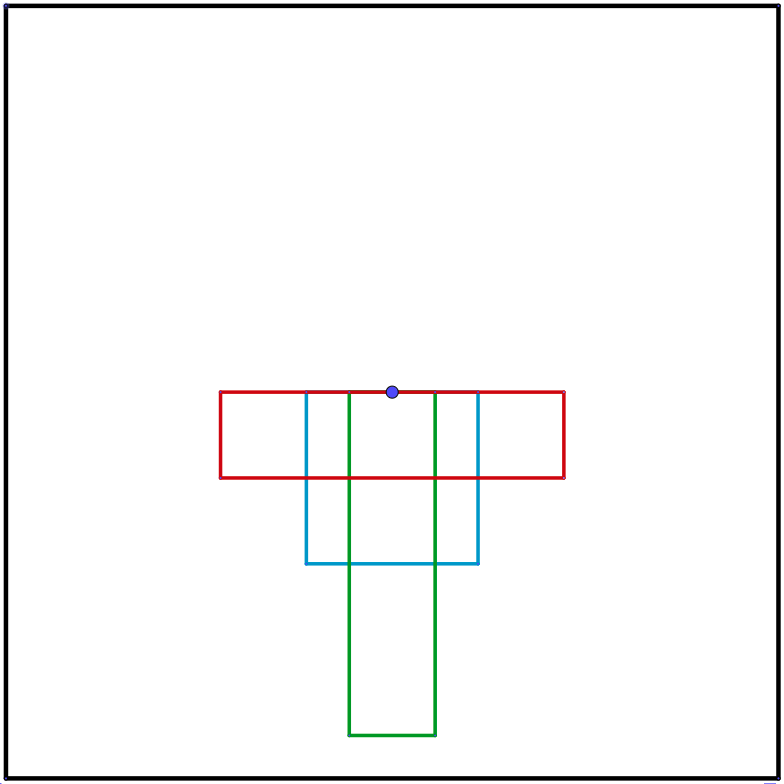}}
    \caption{Radar point-based Anchors. From left to right, anchor boxes towards the center, right, left, bottom, and top edge center of radar point(blue dot).}
    \label{fig3}
\end{figure}
\subsection{Radar Point-Based Anchors and RoIAlign Layer}
Radar signals are reliable in providing an object's location even if they are far away hence have a long-range. When radar points are superimposed on images, although radar signals do not point to all object's in the image but incorporate most of the objects and also those in long range. Therefore, it is reasonable to use radar points for generating region proposals.
Another observation is that radar points are not only present towards the center of the surface of the object but also towards the top, bottom, left, and right edges of objects. Thus, in a good proposal scheme, one cannot just consider the radar point at the center of the bounding box but should also consider the radar point on all four edges of the bounding box.

Anchor boxes are generated with 3 ratios (0.5, 1.0, 2.0) at each radar point and with the consideration that each radar point is at the top, down, left, right edge, and at the center(TDLRC) of an anchor box, as shown in Figure \ref{fig3}. In RPN, Radar points are mapped to each output feature map of FPN. Anchors for each radar point are extracted with different scales based on the level of the feature map.

\begin{figure*}  
\centering  
\begin{subfigure}
  \centering  
  \includegraphics[width=178mm, height=33mm]{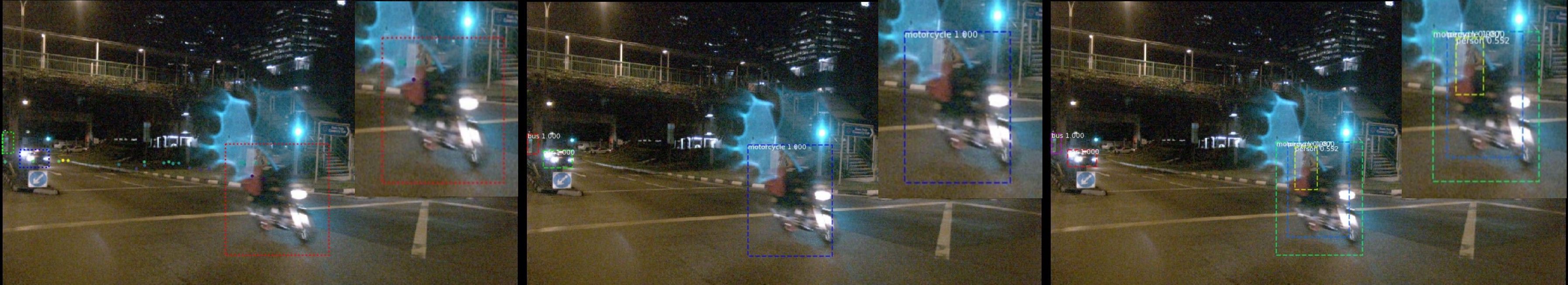}
\end{subfigure}
\begin{subfigure}
  \centering  
  \includegraphics[width=178mm, height=33mm]{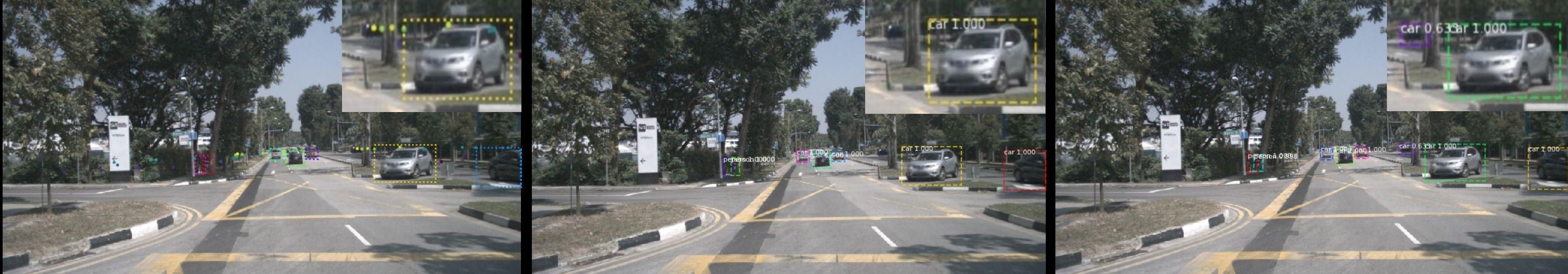}
\end{subfigure}
\caption{Detection sample images. From left to right; Ground-Truth bounding box with Radar point superimposed on Image, FFPN Prediction on RGB, BIRANet Prediction on RGB+Radar.}
\label{fig4}
\end{figure*}

In the region proposal network of FFPN, RoIPooling~\cite{lin2017feature} is used to obtain fixed-size(e.g., 7X7) feature maps from each region of interest. RoIPool, cause harsh quantization over features which creates alignment issue between the input image and extracted features. For a good point based proposals, it is essential to have an accurate mapping of radar points in each feature map. For feature alignment reason, the RoIPool layer of FFPN is replaced with the RoIAlign~\cite{he2017mask} layer.

\subsection{Best of Two - RPN Target Generation Method}
\label{subsec:B2}
In RPN target generation, anchors for each input image are compared to the ground truth bounding boxes, and the IOU score is calculated. Anchors, whose IOU score is greater than 0.7 are assigned with the positive label, and those with IOU score less than 0.3 are assigned with the negative label. Out of all anchor boxes, 128 positive and 128 negative labeled anchor boxes are selected for further process.

In cases where few ground truth bounding boxes of an image do not have any corresponding positive anchor box(i.e., Anchors with IOU$>$0.7), the IOU scores of all corresponding anchor boxes are compared, and the one with maximum IOU score is assigned with a positive label. In such scenarios, radar point-based anchoring offers anchors with higher IOU scores provided there is a radar point corresponding to those ground truth objects. However, it is known that radar sensors do not provide reflected points for all objects in the scene; hence there are no anchors for them.

The Proposed “Best of two” method as outlined in Algorithm \ref{Algo1} help in overcoming this limitation where all anchors as per FFPN along with radar point-based anchors are utilized in RPN target generation. The benefits of this strategy are; first, objects which are not detected by using radar points can now be detected by using RGB anchors similar to FFPN. Second, this method increases the probability of having better anchor candidates for the RPN. The effect of this method can be seen in the performance comparison present in Table \ref{table1} and \ref{table2}. We also want to highlight that the stated method causes a negligible overhead at training time and no increment in inference time.
\begin{algorithm}
 \KwInput{Ground-truth bounding box(gt\_bbox), RGB anchors(i\_anch), radar based anchors(r\_anch).}
 \KwOutput{128 positive target anchors.}
 
 Assign positive/negative labels to RGB anchors and radar based anchors separately.\
 
 \For{positive label RGB anchors(P\_i\_anch)}{
 \For{positive label radar anchors(P\_r\_anch)}
 {
  \If{IOU(P\_r\_anch, gt\_bbox) $>$ IOU(P\_i\_anch, gt\_bbox)}{
   P\_i\_anch = P\_r\_anch.\
   }
 }
 }Select random 128 positive anchors.\
 
 \caption{Best of Two RPN target generation method.}
\label{Algo1}
\end{algorithm}

\section{Dataset}
\label{sec:Dataset}
For training and evaluation, we used the NuScenes dataset which is inspired by the KITTI dataset and covers a large variety of outdoor conditions. To our knowledge, it is the only publicly available dataset where radar data and camera images are synchronized. For object detection, NuScenes provide 3D bounding boxes with 23 classes.

Before using the NuScenes dataset, 3D bounding boxes are converted into 2D and merged relevant classes to obtain 6 classes, which are Car, Truck, Person, Motorcycle, Bicycle, and Bus. In the dataset, there are ground truth bounding box annotations that are not visible in the image due to full occlusion. Such cases are discarded by setting the visibility level parameter from NuScenes as two. Also, the dataset is split into 80:20\% ratio for training and testing, respectively. To investigate the robustness of networks, extra noise is added to evaluation images. Noise added are average blur noise with kernel size three and additive Gaussian noise with 0.05 standard deviation. We also evaluated our networks on extreme amount of noise. Because of page limitations all evaluation and results are not presented here.

\section{Implementation and Training}
\label{sec:implementation}
Faster R-CNN with feature pyramid network is our conceptual base on which our proposed network is built. Experiments are conducted on images of size 1024x1024 and 512x512. Scale used for anchor generation are (16, 32, 64, 128, 256) and (8, 16, 32, 64, 128) respectively. Maximum number of objects per image is set to 100.
Training is conducted on one TITAN Pascal GPU with batch size of two. First, base network FFPN is trained on COCO dataset~\cite{lin2014microsoft} with ResNet50 feature extractor. Then the trained weights are used to initialize the weights of our proposed fused networks(RANet, BIRANet). These network are further trained on the NuScenes training dataset. For training the network, we use step-wise training with 0.001 learning rate.

\section{Results and Evaluation}
\label{sec:eval}
For evaluation, COCO evaluation metric~\cite{lin2014microsoft} is used with an additional AP over 0.85 IOU for a more detailed evaluation. FFPN generates anchors at each pixel of the image, whereas RANet generates anchors only for locations corresponding to which we get radar points. In the presented result Tables \ref{table1}, \ref{table2}, the performance of RANet lack by $\sim$5\% in comparison to that of FFPN, which is a reasonably good performance. Generally, we do not get radar points corresponding to each object in the scene due to the limited field of view, multiple surface reflection, and others. So, it is reasonable to say that a network trained on radar point-based network will not be able to detect objects for which there is no corresponding radar point. BIRANet took care of these missing objects, and provide better detection results in comparison to FFPN. Also, on further observation, we find many cases such as Fig \ref{fig4}, where BIRANet detects objects which are not detected by FFPN.

\begin{table}[htbp]
\resizebox{\columnwidth}{!}{%
\begin{tabular}{l||l|l|l|l|l|l|l|l|l|l|l}
Method Used & AP   & AP50 & AP75   & AP85 & AP(s) & AP(m) & AP(l) & AR & AR(s) & AR(m) & AR(l) \\ \hline
\hline
FFPN(RGB)        & \textbf{69.7}  & 88.2  & 82  & \textbf{60.9}  & 50.3  & 68  & 73.1  & \textbf{73}  & 53.2  & 71.3  & 76.4  \\ \hline
RANet(Radar)     & \textbf{69}    & 83.9    & 80.1   & \textbf{64.4}  & 44.8    & 67.8    & 73.3  & \textbf{71.9}  & 47.3  & 70.9     & 76.2   \\ \hline
BIRANet(RGB+Radar)   & \textbf{72.3}  & 88.9  & 84.3  & \textbf{65.7}  & 53.5  & 70.1  & 76.9  & \textbf{75.3}  & 56.2  & 73.2  & 79.8  \\ \hline
\multicolumn{11}{c}{With Added Noise}                                            \\ \hline
FFPN(RGB)         & \textbf{68.9}  & 88.1  & 81.7  & \textbf{59}  & 49.7  & 67.3  & 72.2  & \textbf{72.1}  & 52.4  & 70.6  & 75.3 \\ \hline
RANet(Radar)       & \textbf{68.3} & 83.3  & 79.1  & \textbf{62.6}   & 45   &66.9    & 73   & \textbf{71.3}    & 47.1  & 70    & 75.9    \\ \hline
BIRANet(RGB+Radar)   & \textbf{71.9}  & {88.9}  & {83.9}  & \textbf{65.1}  & {51}  & {70.2}  & {76.6}  & \textbf{74.4}  & {53.5}  & {73.4}  & {79.5}  \\ \hline
\end{tabular}%
}
\caption{Comparison of detection results on 1024x1024 image size.}
\label{table1}
\end{table}

\begin{table}[htbp]
\resizebox{\columnwidth}{!}{%
\begin{tabular}{l||l|l|l|l|l|l|l|l|l|l|l}
Method Used & AP   & AP50 & AP75   & AP85 & AP(s) & AP(m) & AP(l) & AR & AR(s) & AR(m) & AR(l)  \\ \hline
\hline
FFPN on RGB           & \textbf{65.4} & 87  & 76.8  & \textbf{50.2}  & 44.6  & 63  & 69.9  & \textbf{68.9}  & 48.4  & 66.5  & 73.6  \\ \hline
RANet(Radar)        & \textbf{64.7}  & 82.1  & 75.1 & \textbf{57.4} & 41  & 62.6   & 70.4   & \textbf{67.5}  & 44.1  & 65.5   & 72.9           \\ \hline
BIRANet(RGB+Radar)    & \textbf{68.7}  & 87.6  & 79.7   & \textbf{58.2} & 45.9    & 65.9   & 74.2   & \textbf{72}  & 49.8   & 69.5   & 77.2           \\ \hline
\multicolumn{11}{c}{With Added Noise}                                            \\ \hline
FFPN on RGB  & \textbf{63.7}  & 85.7  & 75  & \textbf{47.9}  & 40.4  & 60.8  & 68.9  & \textbf{67.6}  & 45.2  & 64.7  & 72.8  \\ \hline
RANet(Radar)  & \textbf{63}    & 81.4  & 73.3  & \textbf{53.5}  & 37.4 & 60.5   & 68.6   & \textbf{65.9}       & 40.6  & 63.6  & 71.3    \\ \hline
BIRANet(RGB+Radar)   & \textbf{67.4}  & {87.4}  & {77.7}  & \textbf{56.2}  & {42.7}  & {64}  & {73.8}  & \textbf{70.7}  & {46.8}  & {67.5}  & {76.8}  \\ \hline
\end{tabular}%
}
\caption{Comparison of detection results on 512x512 image size.}
\label{table2}
\end{table}

On further noise addition, we see an expected decrease in performance. However, the drop in performance of BIRANet is less in comparison to that of FFPN.  Both RANet and BIRANet show a minimal drop in performance, which proves their robustness. Results presented in Table \ref{table2} over 512x512 size images, shows that proposed fusion networks are more robust in low-resolution images. Hence, they can be efficient object detectors for vehicles with low-resolution cameras.

Based on the stated findings, we can say that radar data help in making detection more robust towards noise and can perform better. To our knowledge, RRPN is the only work where RGB and radar sensor data of the NuScenes dataset is used in the object detection task. In comparison, our network's performance is found to be better than theirs.

Proposed radar and RGB camera image fusion methods are implemented on FFPN base network. We expect similar behavior when proposed radar and RGB camera image fusion is applied with state of the art object detection networks.

\bibliographystyle{IEEEbib}
\bibliography{refer}

\end{document}